\documentclass[pdflatex,sn-mathphys-num]{sn-jnl}% 

\usepackage{graphicx}%
\usepackage{multirow}%
\usepackage{amsmath,amssymb,amsfonts}%
\usepackage{amsthm}%
\usepackage{mathrsfs}%
\usepackage[title]{appendix}%
\usepackage{xcolor}%
\usepackage{textcomp}%
\usepackage{manyfoot}%
\usepackage{booktabs}%
\usepackage{algorithm}%
\usepackage{algorithmicx}%
\usepackage{algpseudocode}%
\usepackage{listings}%
\usepackage{setspace}
\begin{document}

\title[Article Title]{Training all-mechanical neural networks for task learning through in situ backpropagation}

\author{\fnm{Shuaifeng} \sur{Li}}\email{shuaifel@umich.edu}

\author*{\fnm{Xiaoming} \sur{Mao}}\email{maox@umich.edu}

\affil{\orgdiv{Department of Physics}, \orgname{University of Michigan}, \orgaddress{\city{Ann Arbor}, \postcode{48109}, \state{Michigan}, \country{USA}}}

\abstract{Recent advances unveiled physical neural networks as promising machine learning platforms, offering faster and more energy-efficient information processing. Compared with extensively-studied optical neural networks, the development of mechanical neural networks~(MNNs) remains nascent and faces significant challenges, including heavy computational demands and learning with approximate gradients. Here, we introduce the mechanical analogue of in situ backpropagation to enable highly efficient training of MNNs. We demonstrate that the exact gradient can be obtained locally in MNNs, enabling learning through their immediate vicinity. With the gradient information, we showcase the successful training of MNNs for behavior learning and machine learning tasks, achieving high accuracy in regression and classification. Furthermore, we present the retrainability of MNNs involving task-switching and damage, demonstrating the resilience. Our findings, which integrate the theory for training MNNs and experimental and numerical validations, pave the way for mechanical machine learning hardware and autonomous self-learning material systems.}


\maketitle
\newpage
\section{Introduction}\label{sec1}
The past decades have witnessed the development of artificial intelligence at an unprecedented pace, with machine learning emerging as one of its most transformative branches. At the core of modern machine learning lies neural networks, computational models inspired by the intricate workings of the human brain~\cite{lecun2015deep,jordan2015machine}. Neural networks have revolutionized various fields, ranging from image recognition to natural language processing and autonomous driving~\cite{egmont2002image,adamopoulou2020chatbots,turay2022toward}. Unlike traditional programming, where explicit instructions are provided to solve a problem, neural networks learn from data to make decisions~\cite{abiodun2019comprehensive,carleo2019machine}. This learning process involves adjusting the parameters of interconnected nodes, or neurons, within the network through backpropagation to conduct gradient descent~\cite{amari1993backpropagation,lillicrap2020backpropagation}. Eventually, neural networks can uncover complex patterns and relationships in data, enabling them to generalize to unseen examples and perform tasks with remarkable accuracy.

Nevertheless, the substantial computational requirements and energy consumption associated with computer-based neural networks, especially considering the energy efficiency of conventional digital processors, present significant challenges to further development~\cite{thompson2020computational,sze2017efficient}. One proposed solution lies in physical machine learning hardware platforms, such as optical and mechanical neural networks, which hold promise for greater speed and energy efficiency compared to their digital counterparts~\cite{hamerly2019large,caulfield2010future,hughes2019wave,wetzstein2020inference,shastri2021photonics,wang2022optical,pai2023experimentally,hermans2015trainable,wright2022deep,jiang2023metamaterial}. For example, optical neural networks, extensively studied, boast an energy advantage of several orders of magnitude over electronic processors utilizing digital multipliers~\cite{wang2022optical}. Besides, the efficient and feasible training method of in situ backpropagation for optical neural networks promotes the promising physical machine learning platform and the reduction of carbon footprint~\cite{hughes2019forward,hughes2019wave,pai2023experimentally}.

Wave-matter interactions are commonly utilized to implement machine learning in optical neural networks~\cite{lin2018all,hughes2019wave} and MNNs~\cite{jiang2023metamaterial}, mechanisms including the diffraction and the equivalence between recurrent neural networks and wave physics, but using wave dynamics may encounter challenges such as energy dissipation and complex implementations in real-world applications. In contrast, the physical change induced in MNNs under static forces offer a promising solution to address these challenges. Moreover, MNNs exhibit superiority over optical counterparts in some extreme cases, including complex electromagnetic environments, yet they remain relatively unexplored in terms of efficient training methods and experimental demonstrations. For instance, while training MNNs to learn behaviors has been experimentally demonstrated, the approach using optimization algorithms~(e.g., genetic algorithms) operated on computers ultimately relies on conventional digital processors~\cite{lee2022mechanical,hopkins2023using}. 

In light of this, the concept of \emph{physical learning} offers a promising avenue to train MNNs only using local information of the networks without the computer aid~\cite{stern2020supervised,stern2021supervised,stern2023learning,altman2023experimental}. While effective across various machine learning tasks, the coupled learning rule, inspired by contrastive Hebbian learning, by considering the response in two steady states of the system, yields the approximate gradient of the loss function rather than the exact one~\cite{stern2020supervised}. 

Here, we present a highly-efficient training protocol for MNNs through mechanical analogue of in situ backpropagation, derived from the adjoint variable method, in which the learning only relies on the local information. By using 3D-printed MNNs, we demonstrate the feasibility of obtaining the exact gradient of the loss function experimentally solely from the bond elongation of MNNs in only two steps, essentially without a computer.  Besides, leveraging the obtained gradient, we showcase the successful training of a mechanical network for behaviors learning and various machine learning tasks, achieving high accuracy in both regression and Iris flower classification tasks, validated both numerically and experimentally. In addition, we illustrate the retrainability of MNNs after switching tasks and damage, a feature that may inspire further inquiry into more robust and resilient design of MNNs.

Beyond their applications as computational devices, these MNNs also offer unprecedented opportunities in materials science and mechanical engineering as sustainable and autonomous materials systems.  In engineering, few examples exist where materials or machines possess the innate ability to exhibit desired behaviors without meticulous design and engineering. However, the design strategies need expert knowledge and experience. 
The MNNs as well as the efficient training protocol proposed here also pave the way for future intelligent material systems which can be repeatedly trained to adapt to different environments and tasks.

\section{Results}\label{sec2}
\subsection{In situ backpropagation in mechanical neural networks}\label{sec2_1}
We start from introducing the theoretical basis to conduct in situ backpropagation in MNNs. For a MNN without zero modes under boundary conditions, owning $n$ nodes and $m$ springs in $d$ dimension, given the certain task, the task learning problem can be described as:
\begin{equation}
    \label{eqn1}
    \begin{aligned}
        & \underset{k}{\text{minimize}} & & \mathcal{L}[u(k)],\\
        & \text{subject to} & & Du=F,
    \end{aligned}
\end{equation}
where $\mathcal{L}$ is the loss function, $k\in\mathbb{R}_{\geq0}^{m\times 1}$ is a vector containing the spring constant of each bond, which is the trainable learning degree of freedom, $D\in\mathbb{R}^{dn\times dn}$ is the stiffness matrix, $u\in\mathbb{R}^{dn\times1}$ is the node displacement, which is the output and $F\in\mathbb{R}^{dn\times1}$ is the external forces applied on the nodes, which is the input. The governing equation of statics, $Du=F$, represents the forward problem, reflecting the response $u$~(displacement of each node) of a MNN under input forces $F$. To minimize $\mathcal{L}[u(k)]$ using gradient descent, $\nabla\mathcal{L}$ is derived as below:
\begin{equation}
    \label{equ2}    
    \nabla\mathcal{L}=\frac{d\mathcal{L}}{dk}=\frac{\partial \mathcal{L}}{\partial u}\frac{du}{dk},
\end{equation}
Given the form of the loss $\mathcal{L}$ as a function of $u$, Jacobian $\frac{\partial\mathcal{L}}{\partial u}$ can be conveniently calculated, whereas usually $\frac{du}{dk}$ is a computationally-heavy term due to interactions between the nodes. We show as below that this term can be derived from the differentiation of $Du=F$ on both sides:
\begin{equation}
    \label{equ3}
    \frac{du}{dk}=-D^{-1}\frac{dD}{dk}u.
\end{equation}
Plug Eq.~\eqref{equ3} to Eq.~\eqref{equ2}:
\begin{equation}
    \label{equ4}
    \nabla\mathcal{L}=\frac{\partial \mathcal{L}}{\partial u}\left(-D^{-1}\frac{dD}{dk}u\right)
    =u^{*}\frac{dD}{dk}u.
\end{equation}
Here we use $u^{*}$ to represent $-\frac{\partial \mathcal{L}}{\partial u}D^{-1}$ to obtain $Du^{*T}=-\left(\frac{\partial \mathcal{L}}{\partial u}\right)^{T}$. Therefore, after defining $u^{*T}$ as adjoint displacement field $u_{adj}$, the adjoint problem can be expressed as below:
\begin{equation}
    \label{equ5}
    Du_{adj}=-\left(\frac{\partial \mathcal{L}}{\partial u}\right)^{T}.
\end{equation}
Then, in the linear regime, after introducing the compatibility matrix $C\in\mathbb{R}^{m\times dn}$ that maps the node displacement $u$ to the bond elongation $e$, the gradient of $\mathcal{L}$ can be expressed as:
\begin{equation}
    \label{equ6}
    \nabla\mathcal{L}=u^{*}\frac{dD}{dk}u=u^{*}\frac{d(C^{T}KC)}{dk}u=u_{adj}^{T}C^{T}\frac{dK}{dk}Cu=e_{adj}\circ e,
\end{equation}
where $K$ is the diagonal matrix with $k$ as the diagonal entries and $\circ$ is the Hadamard product~(i.e., element-wise product). $\frac{dK}{dk}$ is a tensor $\delta_{pql}\in\mathbb{R}^{m\times m\times m}$, where the entry is $1$ when $p=q=l$ and otherwise $0$. Eq.~\eqref{equ6} implies that the gradient of the loss function $\mathcal{L}$ equals to the element-wise multiplication of elongations of the bonds in the forward problem $Du=F$ and the adjoint problem $Du_{adj}=-\left(\frac{\partial \mathcal{L}}{\partial u}\right)^{T}$.

Apparently, two problems only have the difference on the input force. Therefore, to implement in situ backpropagation in MNNs and obtain the gradient of the loss function $\mathcal{L}$ from the local information of MNNs, there are two steps: ($1$) Apply the input force $F$ to the MNNs, and obtain the displacement of the nodes and forward elongation of the bonds $e$. ($2$) Calculate $-\left(\frac{\partial \mathcal{L}}{\partial u}\right)$ given the form of the loss function $\mathcal{L}(u)$ using the displacement in step ($1$) and apply the force $-\left(\frac{\partial \mathcal{L}}{\partial u}\right)^{T}$ to the same system to obtain the adjoint elongation of the bonds $e_{adj}$. The gradient is the element-wise multiplication of the forward elongation and the adjoint elongation. Note that entries in $-\left(\frac{\partial \mathcal{L}}{\partial u}\right)$ are only nonzero at the output nodes. Therefore, essentially the backward signals propagate from output nodes, similar to the backpropagation in computer-based and optical neural networks~\cite{lillicrap2020backpropagation,hughes2019forward,hughes2019wave,hughes2018training}. 

More importantly, this method for in situ backpropagation stays consistent with the local rule required in physical learning~\cite{stern2021supervised}, since the gradient for bond $i$ can be obtained solely from the elongation of bond $i$, i.e., $\nabla\mathcal{L}_{i}=e_{adj,i}e_{i}$. Different from the local learning rule which relies on approximated gradient, our method shows the advantage of obtaining exact gradient, which might be able to avoid the unexpected spikes in the decreasing loss when using coupled learning rules~\cite{dillavou2022demonstration,wycoff2022desynchronous}.

Subsequently, this gradient $\nabla\mathcal{L}_{i}$, obtained locally at all bonds $i$ via these two steps described above, is used to update the spring constants at learning rate $\alpha$, from $k_i$ to $k_{i} - \alpha \nabla\mathcal{L}_{i}$
\begin{equation}
    \label{equ7}
    k_{i} \leftarrow k_{i} - \alpha \nabla\mathcal{L}_{i}
    = k_{i} - \alpha e_{adj, i}e_{i},
\end{equation}
iteratively through gradient descent, minimizing the loss function subject to the physics law.

Here, to demonstrate the in situ backpropagation in MNNs, we fabricate two-dimensional MNNs made of flexible Agilus30 using 3D printing techniques, as shown in Fig.~\ref{fig1}a. The detailed fabrication procedures and configurations are shown in Methods and Supplementary Information. As an example, we take the loss function to be $\mathcal{L}=(u_{Ly}+0.025~\mathrm{m})^2$ under the downward applied force $F_{Ry}=0.01\times9.8~\mathrm{N}$, where $u_{Ly}$ and $F_{Ry}$ represent the vertical displacement of the node at bottom left and the applied force on the node at bottom right of MNNs. The leftmost and rightmost sides at the top of our MNN are glued on the truss as the fixed boundary condition. The force is applied by the weights through gravity.

Fig.~\ref{fig1}b shows the experimental setup to implement the mechanical analogue of in situ backpropagation. The first panel shows the forward field where $F_{Ry}$ is applied on the bottom right node by a $10~\mathrm{g}$ weight. The first panel of Fig.~\ref{fig1}c exhibits the experimentally measured elongation of each bond, and the corresponding simulated one is shown in the first panel of Fig.~\ref{fig1}d. The applied force results in the vertical displacement of the node on the bottom left, which is measured to be $u_{Ly}=-0.82~\mathrm{mm}$ experimentally. Therefore, the value of the nonzero entry of the force for the adjoint problem $-2(u_{Ly}+0.025)$ is calculated to be equivalent to $5~\mathrm{g}$ weight hung on the bottom left node, with experimental setup displayed in the second panels of Fig.~\ref{fig1}b. Note that the calculated force is converted to weights used in experiments by gravitational acceleration constant $g=9.8~\mathrm{m/s^2}$. Figs.~\ref{fig1}c and~\ref{fig1}d show the measured and simulated elongation, respectively. Our method indicates that the gradient of the loss function is the element-wise multiplication of the forward elongation and the adjoint elongation, which is illustrated in the third panels of Figs.~\ref{fig1}c and~\ref{fig1}d, representing experimental and simulated results, respectively.

We observe that our measured elongation and simulated elongation have an excellent agreement, as well as the gradient. Compared with the simulated gradient, which represents the exact gradient, our experimental gradient achieves over $90\%$ accuracy, as shown in the inset in Fig.~\ref{fig1}e. These results are summarized by three independent experiments described in the main text, along with additional three independent experiments for another loss function, which is detailed in the Supplementary Information.

To provide a comparison, we calculate the approximate gradient using the forward difference as follows:
\begin{equation}
    \label{equ8}
    \nabla\mathcal{L}=\frac{\mathcal{L}[u(k+\delta k)]-\mathcal{L}[u(k)]}{\delta k}.
\end{equation}
The error depends on the step size $\delta k$, where the smallest error occurs at around $\delta k=10^{-6}~\mathrm{N/m}$, as shown in the left panel of Fig.~\ref{fig1}e. When the step size is smaller than the optimal one, the fixed number of binary digits in computers leads to the roundoff error, and the truncation error will emerge when using larger step size. In sharp contrast, the numerical implementation of our method does not produce error. Besides, finite difference method is usually implemented numerically since $\delta k$ needs to be considerably small~(also see the experimentally feasible step size within the shaded area in Fig.~\ref{fig1}e with large numerical error), whereas our method features experimental feasibility by only measuring the bond elongation. Furthermore, as shown in the right panel of Fig.~\ref{fig1}e, for the finite difference method, the number of required simulations increases linearly as the number of bonds in networks increases because each element $\partial \mathcal{L}/\partial k_i$ corresponding to bond $i$ requires a separate computation, whereas our adjoint method only needs two simulations or experiments to obtain the gradient regardless of the number of bonds in networks.

In addition, another advantage of our MNNs is the potential utilization of the same node as both input and output node, owing to the noncontroversial nature of input force and output displacement. This characteristic enables more compact design of MNNs. In the Supplementary Information, we demonstrate the in situ backpropagation when the defined loss function and the input force are in the same node. Furthermore, to check if our bar model used for simulation agrees with the actual experimental samples, as part of the supplementary analysis for simulations, we use the finite element method to analyze 3D actual experimental samples, which yields results in accordance with the predominantly used bar model~(see Supplementary Information for further details).

\subsection{Behaviors learning}\label{sec2_2}
As mentioned in the Introduction, training MNNs to learn behaviors can reduce the effort of design strategies. Here, we show that without expert knowledge, through in situ backpropagation MNNs can learn desired behaviors. For example, for a MNN with uniform bonds without deliberate design, as shown in Fig.~\ref{fig2}a, when the input force $F=0.005\times 9.8~\mathrm{N}$ is applied on the red node, the two cyan nodes will have the same vertical displacements $u_{Ly}=u_{Ry}$~(i.e., symmetric output) due to the symmetric configuration, where subscript $L$ and $R$ represent the node on the left and right, respectively. 

Considering two classes represented by two cyan nodes, we can use the cross-entropy loss with the normalization~\cite{good1956some}, $\mathcal{L}=-\sum_{c=1}^{N}y_{c}\ln{p_{c}}$, where $N=2$, $y_{c}$ and $p_{c}=\frac{e^{|u_{c}|}}{\sum_{c=1}^{N}e^{|u_{c}|}}$ represent the number of classes ($2$ for $L$ and $R$), binary indicator and predicted probability of $u_{c}$, respectively, to maximize the probability of the vertical displacement of one of the nodes. For example, $\{y_1,y_2\}=\{1,0\}$ for class 1, where the left node has greater displacement, and vice versa. The cross-entropy loss decreases as the predicted probability $p_{c}$ approaches the actual label, leading to the maximization of the probability and the difference between two absolute vertical displacements. Through in situ backpropagation, the asymmetric output can be realized, where two nodes have different vertical displacement under the same force applied on the red node. Fig.~\ref{fig2}b shows the progressive reduction in loss during training until convergence. Meanwhile, the difference of absolute vertical displacements between two nodes increases until reaching the maximum. The training process including loss decrease and bonds change is also shown in Supplementary Video $1$.

Fig.~\ref{fig2}c depicts the trained MNN, featuring a node on the left with a larger displacement in response to the input force. On the contrary, in Fig.~\ref{fig2}d, another trained MNN presents an alternate scenario where the node on the right displays a greater displacement under the same input force. Notably, experimental measurements align closely with simulated displacements across all scenarios, validating the efficacy of the trained MNNs. It is noteworthy that using the cross-entropy function as the loss encourages the maximization of the displacement difference between two nodes, rather than a specific value when using mean-squared error~(MSE) as the loss. In addition, we demonstrate the precise control of displacements of two nodes under the applied force $F=0.005\times 9.8~\mathrm{N}$ by using MSE, as detailed in the Supplementary Information. The utilization of in situ backpropagation offers a straightforward methodology for training mechanical structures to exhibit desired functionalities, thereby paving the way for future applications in the design of intricate mechanical systems, including automotive design and robotics.

\subsection{Machine learning}\label{sec2_3}
As outlined in the Introduction, akin to their computer-based counterparts, MNNs offer a compelling avenue for implementing machine learning tasks with enhanced speed and energy efficiency. Furthermore, the model for a machine learning task is essentially stored in a MNN by the real materials, showing interpretability. In this section, we select two representative tasks typically undertaken by computer-based neural networks to showcase the versatility and efficacy of MNNs.

\subsubsection{Training mechanical neural networks for regression tasks}\label{sec2_3_1}
Regression stands as benchmark tasks in the field of machine learning, serving as a cornerstone for evaluating model performance and predictive capabilities. Here, given that the in situ backpropagation in MNNs is conducted within the linear regime, linear regression is chosen to demonstrate in our case. Considering the stiffness of our experimental MNNs, we choose four synthetic datasets to exemplify the regression task as a function of the input force $F$, expressed as follows:
\begin{equation}
    \label{equ9}
    \begin{aligned}
        & u_{Rx}=0F,\\
        & u_{Ry}=0.016F,\\
        & u_{Lx}=0.004F,\\
        & u_{Ly}=0.016F.
    \end{aligned}
\end{equation}
Here, $u_{Rx}$, $u_{Ry}$ are the horizontal and vertical displacement of the bottom right node, respectively. $u_{Lx}$ and $u_{Ly}$ are those of the bottom left node. Hence, by training with such dataset, we expect these two nodes to exhibit linear displacement under applied force according to the prescribed slopes. This regression task can also be simply visualized by two straight lines as shown in Fig.~\ref{fig3}a, formulated as follows:
\begin{equation}
    \label{equ10}
    \begin{aligned}
        & u_{Rx}=0,\\
        & u_{Ly}=4u_{Lx}.
    \end{aligned}
\end{equation}
This can be interpreted as relations between the horizontal and vertical displacements of two nodes, where the bottom right node does not develop horizontal displacement while the bottom left node has horizontal and vertical displacements in a certain relation under the applied force. We generate a set of $100$ random data points according to Equation~\ref{equ9}, as depicted in the left penal of Fig.~\ref{fig3}a. In addition, the noisy data are also generated from Eq.~\eqref{equ9} by adding the Gaussian noise with a standard deviation of $10^{-4}$, as illustrated in the right panel of Fig.~\ref{fig3}a.

We randomly split the synthetic dataset into a training set~($70\%$) and a testing set~($30\%$). As displayed in Fig.~\ref{fig3}b, the MSE losses~($\mathcal{L}=\frac{1}{N}\sum_{j=1}^{N}(u_{j}-\hat{u}_{j})^{2}$, where $N=100$, $u_{j}$ and $\hat{u}_{j}$ are predicted values from current MNNs and target values in the synthetic dataset, which are composed of horizontal and vertical displacements, respectively.) for the noise-free and noisy datasets exhibit consistent decrease over epoch until convergence. This decline indicates an improved fit between regression results and datasets. Fig.~\ref{fig3}c, from left to right, illustrates the regression results under different epoch. At the beginning of training~($epoch=1$), a cross, which has large discrepancy from Fig.~\ref{fig3}a, is shown. As the epoch increase to $10$, the orange line gradually becomes vertical and blue line remains tilted. Upon convergence of the loss, the regression results at $epoch=5000$ show the excellent agreement with regression targets. The training process including loss decrease, bonds change and relevant regression results is shown in Supplementary Video $2$.

Fig.~\ref{fig3}d exhibits the trained MNN, where the widths of bonds are different. Therefore, it is anticipated that upon applying different force to the red node, the nodes marked by the orange and blue stars will develop displacement in accordance with the solid lines depicted in Fig.~\ref{fig3}a. Fig.~\ref{fig3}e shows the experimental setup under the applied force $F=0.006\times9.8~\mathrm{N}$. The experimentally measured displacement $u_{x}$ and $u_{y}$ of two nodes L and R, marked by stars, under the applied force ranging from $F=0\times9.8~\mathrm{N}$ to $F=0.012\times9.8~\mathrm{N}$ with an increment $\Delta F=0.002\times9.8~\mathrm{N}$ are presented in the third panel of Fig.~\ref{fig3}c, where the good agreement with simulations can be observed. 

This regression task can also be interpreted as a behaviors learning task, where the trajectory of the node under the applied force can be precisely engineered using in situ backpropagation. The behaviors learning task using regression poses a greater challenge than the traditional behaviors learning task, as the desired outputs are not specific values, but instead, functions of the input force. In our specific case, while the bottom right node remains stationary horizontally under downward force, the bottom left node demonstrates both horizontal and vertical displacements in a proportionate manner. Leveraging regression to implement behaviors learning opens up new avenues for designing materials functionalities.

\subsubsection{Training mechanical neural networks for classification tasks}\label{sec2_3_2}
Another benchmark task in machine learning is classification. In our study, we utilize the well-known Iris flower dataset, a real-world dataset, to exemplify the classification process~\cite{iris_53}. This dataset aims to classify three types of Iris flower -- namely, Iris setosa, Iris versicolor and Iris virginica -- using four distinct features: sepal length, sepal width, petal length, and petal width. Fig.~\ref{fig4}a visually illustrates the relation between sepal length and petal length among three species. These species exhibit clear boundaries in the feature space, with Iris virginica typically characterized by larger sepal and petal lengths compared with the other species.

In our classification process, we consider four features, each corresponding to a downward input force applied on the nodes marked by red dots in the inset of Fig.~\ref{fig4}b. Note that these four values are appropriately scaled based on the stiffness of experimental MNNs and details can be seen in Supplementary Information. The indicator of a specific species is determined by the node with the largest horizontal displacement among the three nodes in the inset marked by the corresponding symbols.

In our classification task, we employ the cross-entropy loss function and randomly partition the dataset into a training set~($70\%$) and a testing set~($30\%$). As the loss steadily decreases over epoch, the accuracy of the classification defined by the ratio of the correct classification for the training dataset approaches nearly $100\%$. Meanwhile, the accuracy for the testing dataset, which is unseen during the training process, also converges to nearly $100\%$, suggesting that our MNNs have effectively learned the complex pattern and relations in this dataset. The trained MNNs is shown in the inset with different width of the bonds. Fig.~\ref{fig4}c illustrates the classification results at different epoch. From left to right, with the increase of epoch, the Iris setosa is classified first as it is distinctly separated from the other species in the feature space. Gradually, Iris virginica emerges, eventually sharing a clear boundary with Iris versicolor in the feature space. The Supplementary Video $3$ shows the clear evolution of classification results as the training goes on. Note that the third panel of Fig.~\ref{fig4}c exhibits remarkable similarities with Fig.~\ref{fig4}a which is the ground truth, affirming the efficacy of our classification model. The confusion matrix can be seen in the Supplementary Information, showing the classification results from MNNs are becoming closer to the ground truth over epoch.

Furthermore, we perform classification experiments to validate the model stored in MNNs. The input forces corresponding to each Iris flower are rounded and converted to integer weights, which are then applied to the corresponding nodes, as shown in the insets of Fig.~\ref{fig4}d. The details can be seen in the Supplementary Information. Fig.~\ref{fig4}d shows, from top to bottom, the normalized measured horizontal displacement using features of Iris virginica, Iris versicolor and Iris setosa, respectively. Remarkablely, in both simulation and experiment, the largest horizontal displacement occurs at the corresponding node, indicating correct classifications. This consistency between simulation and experimental results reinforces the reliability of our classification model.

\section{Retrainability}\label{sec3}
In the preceding sections, we have showcased the remarkable capability of our MNNs to implement the behaviors learning and various machine learning tasks through in situ backpropagation. Distinct from the computer-based neural networks, which exist solely in the digital realm, MNNs are physically manufactured, embedding the machine learning model within real materials. Hence, the retrainability of MNNs emerges as a pivotal attribute~\cite{dillavou2022demonstration}. Here, we highlight the retrainability in two key scenarios: first, their ability to seamlessly transition from one task to another on demand, and second, their capacity to recover the machine learning model after sustaining damage.

In the task-switching scenario, we start from the Iris flower classification task. Following training our MNN until convergence as shown in Fig.~\ref{fig5}a, the resulting trained MNN for the classification task serves as the initial configuration for the regression task. We notice that in Fig.~\ref{fig5}b the mean-squared error decreases and the accuracies for both the training and testing noise-free dataset approach nearly $100\%$, with accuracy defined as the $l^{2}$-norm error. Fig.~\ref{fig5}c shows the decreasing cross-entropy loss and increasing accuracy as the task switches to classification. Upon convergence, the trained MNN differs from that depicted in Fig.~\ref{fig5}a, suggesting the different local minima when starting from distinct initial configurations. Although the accuracies in both Fig.~\ref{fig5}a and Fig.~\ref{fig5}c eventually increase to nearly $100\%$, yet, the converged loss in Fig.~\ref{fig5}a is smaller than that in Fig.~\ref{fig5}c, indicating a more pronounced classification sign~(i.e., larger horizontal displacement of the corresponding node compared with that of other two nodes in our case). The Supplementary Video $4$ depicts the transition process from classification to regression and back to classification.

Another scenario involves the retrainability of the MNNs after damage occurs. We again begin with the Iris flower classification task shown in Fig.~\ref{fig5}a. Then, we prune one of the bonds in the trained MNN displayed in Fig.~\ref{fig5}d, effectively resulting in the breakdown of the Iris flower classification model stored in our MNNs. As shown in Fig.~\ref{fig5}e, the classification accuracy diminishes to approximately $50\%$, accompanied by an increase in the cross-entropy loss, signifying the degradation of the classification model. However, after training the damaged MNNs, the classification accuracy rebounds to around $80\%$, indicating a substantial recovery of the classification model stored in MNNs. It is worth noting that the decrease in loss exhibits relatively large variation, suggesting that under such configuration the training process depends on the partitioning of training and testing datasets. The Supplementary Video $5$ illustrates the retrainability of the MNNs following damage occurrence.

While the pruning of one bond is demonstrated above, our Supplementary Information delves into the effects of pruning different bonds on classification accuracy, where distinct bonds exhibit varying importance in classification tasks. Pruning ``redundant'' bonds sustains high accuracy levels; however, pruning ``critical'' bonds causes a significant accuracy decline. Moreover, the pruning of certain bonds can render the MNN mechanically unstable, leading to the emergence of zero modes. The identification of ``critical'' and ``redundant'' bonds in the classification task, as demonstrated in the Supplementary Information, underscores the robustness and vulnerability of our MNNs. It stimulates the development of more resilient MNN designs, encompassing network topology and connectivity considerations~\cite{beygelzimer2005improving,dekker2004network}. Furthermore, the examination of damaged MNNs may prompt reflection on their potential parallels with damaged biological neural networks in the brain, thereby inspiring further exploration into shared characteristics~\cite{kalampokis2003robustness,eluyode2013comparative}.

\section{Conclusion}\label{sec4}
In conclusion, our study presents a foundational method for training MNNs through in situ backpropagation, derived from the adjoint variable method. This novel approach enables the computation of gradients of the loss function from local information within MNNs in only two steps, demonstrating remarkable efficiency. Leveraging in situ backpropagation, we have investigated the capabilities of MNNs in learning behaviors and various machine learning tasks, achieving high accuracy in both regression and Iris flower classification tasks. These physically-manufactured MNNs store machine learning models in real materials, distinguishing them from the computer-based neural networks.

Moreover, our work highlights the retrainability of MNNs, a crucial attribute with profound implications for real-world applications. We have demonstrated that these networks can seamlessly switch between tasks and recover from damage, showcasing their robustness and resilience. In contrast to other physical neural networks, such as optical neural networks, our MNNs offer greater ease of operation in experiments and real-world applications across diverse environments. The utilization of static force to implement task learning addresses some challenges in physical neural networks based on wave dynamics, such as energy dissipation and loss. Also, MNNs provide fairly fast information processing at the speed of sound of materials in both in situ backpropagation at the training stage and decision-making at the prediction stage.

\begin{figure}[h]
    \centering
    \includegraphics[width=0.9\textwidth]{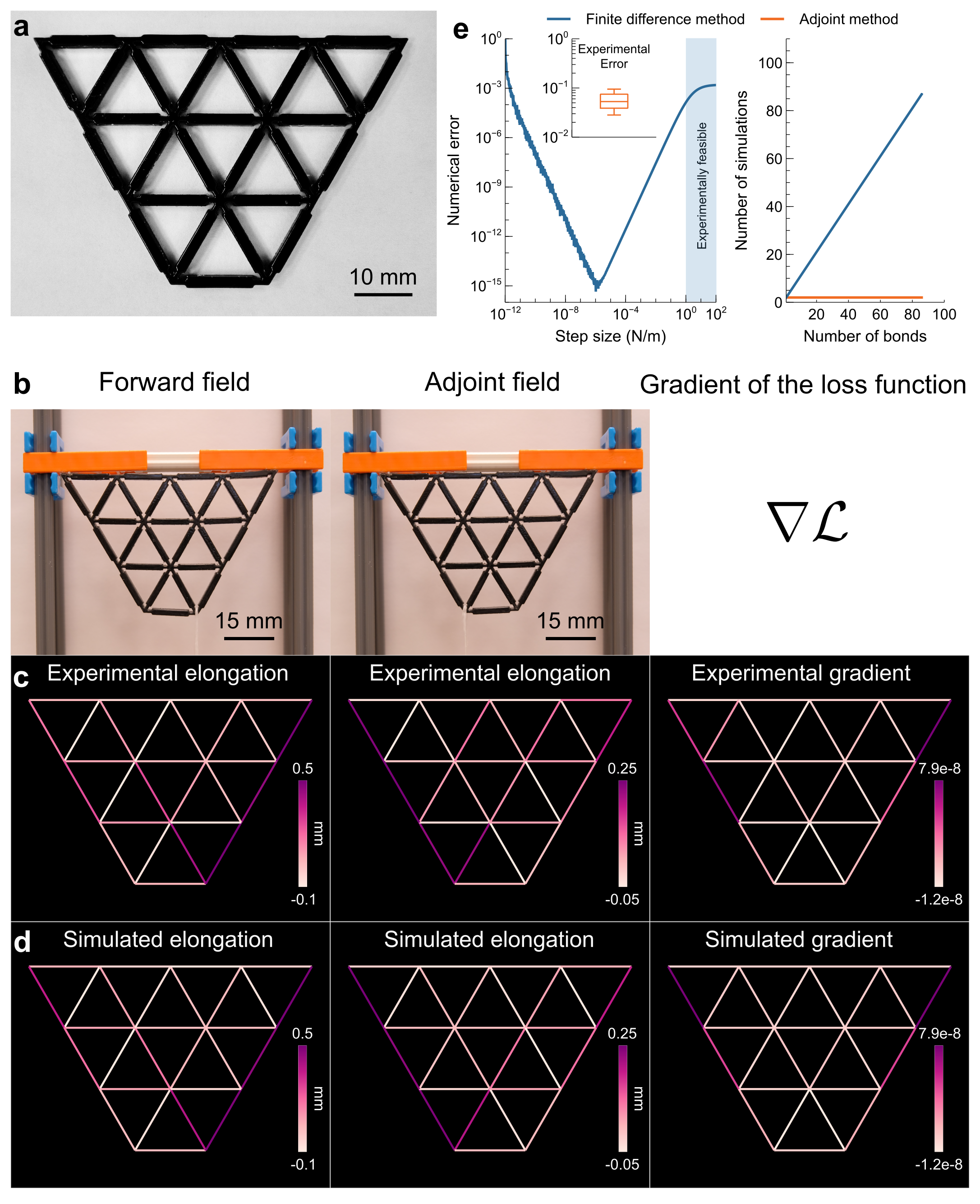}
    \caption{\textbf{Experimental demonstration of the in situ backpropagation.}
    \textbf{a} 3D-printed mechanical networks using the Polyjet rubber-like material Agilus30.
    \textbf{b} The experimental setups for the forward field and the adjoint field, and the resulting gradient of the loss function are shown from left to right.
    \textbf{c} The experimental elongations of forward field and adjoint field are shown in the first and second panel, respectively. The experimental gradient is shown in the third panel.
    \textbf{d} The corresponding simulated elongations and gradient are shown from left to right.
    \textbf{e} The comparison between the finite difference method and our adjoint method. The left panel shows the numerical error as a function of step size in finite difference method. The shaded area represents the experimentally feasible region with large step sizes, below which the step size $\Delta k$ is too small for manufacturing accuracy. The inset shows the experimental error in adjoint method. The numerical error of the adjoint method is zero. The right panel shows the number of required simulations of the finite difference method and the adjoint method as a function of the number of bonds in the MNNs to obtain the gradient.
    }
    \label{fig1}
\end{figure}

\begin{figure}[h]
    \centering
    \includegraphics[width=0.9\textwidth]{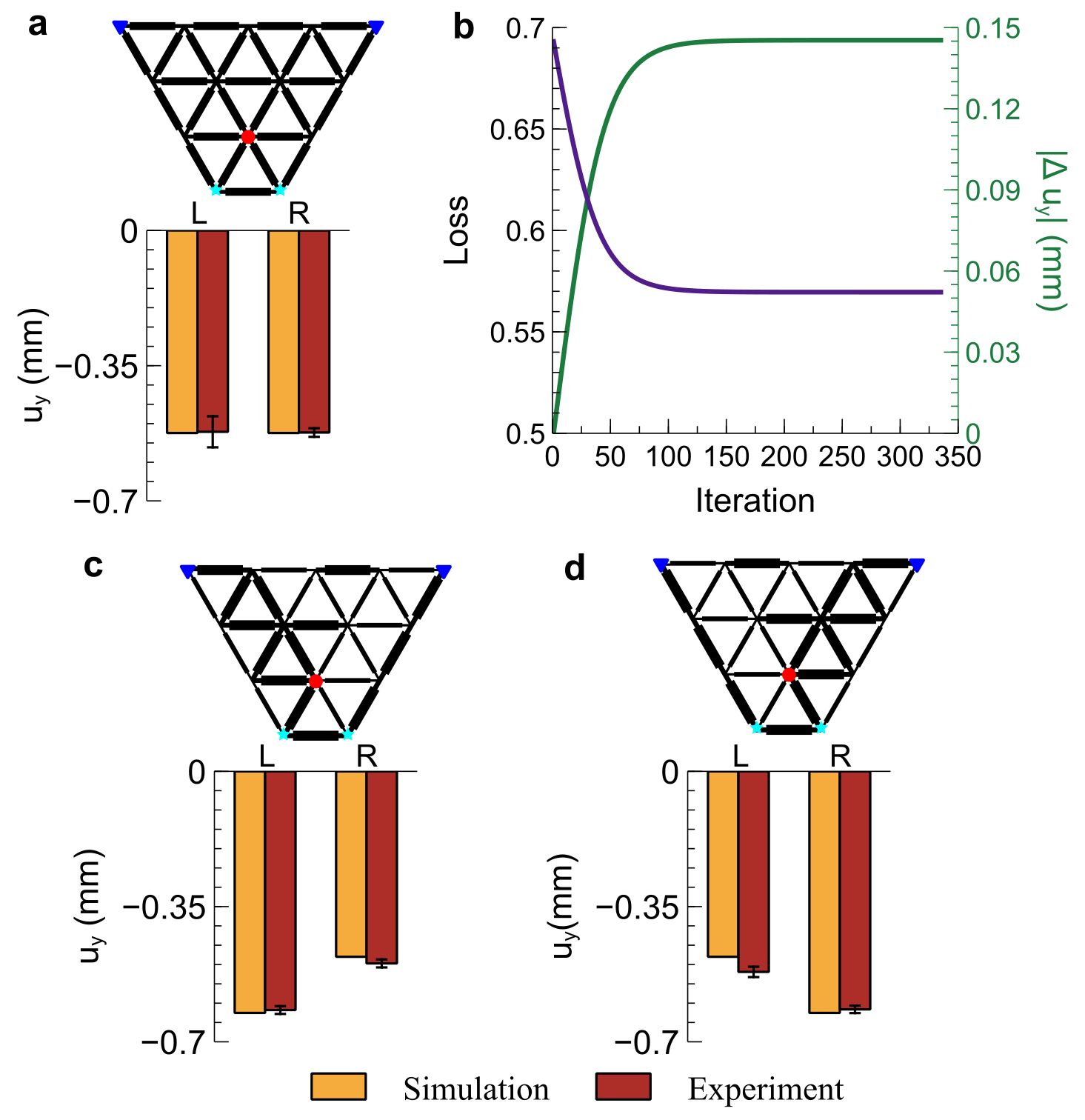}
    \caption{\textbf{Behaviors learning using MNNs.}
    \textbf{a} The symmetric output under the applied force of the MNN before training. The top panel shows the configuration of mechanical networks. The bottom panel shows the simulated and experimental vertical displacements $u_{y}$ of two nodes.
    \textbf{b} The loss and the absolute difference of vertical displacements of two nodes $|\Delta u_{y}|$ as a function of iteration in the training process.
    \textbf{c, d} Two asymmetric outputs under the applied force as a result of the training.
    The top panels in \textbf{c} and \textbf{d} show the configuration of MNNs. The bottom panels in \textbf{c} and \textbf{d} show the simulated and experimental vertical displacements $u_{y}$ of two nodes.
    The blue triangles, red dots and cyan stars in the top panels of \textbf{a}, \textbf{c} and \textbf{d} represent the fixed nodes, the input node and the output nodes, respectively.
    The error bars in the bottom panels of \textbf{a}, \textbf{c} and \textbf{d} are calculated based on standard deviation of three independent experiments.
    }
    \label{fig2}
\end{figure}

\begin{figure}[h]
    \centering
    \includegraphics[width=0.9\textwidth]{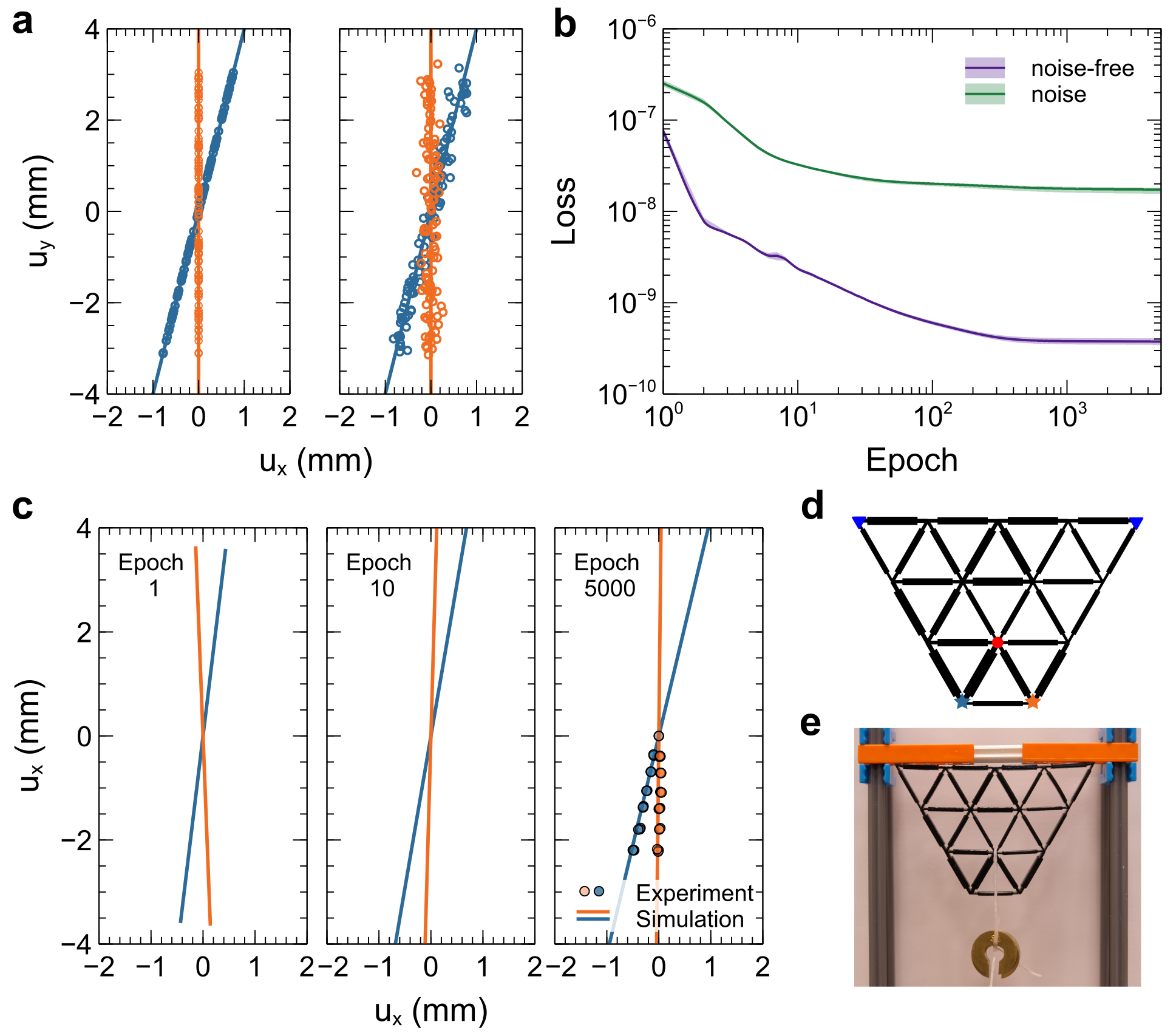}
    \caption{\textbf{Linear regression using MNNs.}
    \textbf{a} The synthetic noise-free and noisy dataset~(circles) are exhibited from left to right. The regression results are shown in solid lines.
    \textbf{b} The loss for the noise-free dataset~(purple) and noise dataset~(green) as a function of epoch in training process.
    \textbf{c} The simulated regression results~(solid lines) when the epoch is $1$, $10$ and $5000$ are shown from left to right, respectively. The experimental regression results are exhibited in circles when conducted by the MNN at epoch $5000$. Note that experimental results of three independent experiments are almost overlapped.
    \textbf{d} The trained configuration of MNNs for regression tasks. The blue triangles, red dot and stars represent the fixed boundary, the input node and the output nodes, respectively.
    \textbf{e} The experimental setup for the regression task when the input force is equivalent to $6~\mathrm{g}$.
    }
    \label{fig3}
\end{figure}

\begin{figure}[h]
    \centering
    \includegraphics[width=0.9\textwidth]{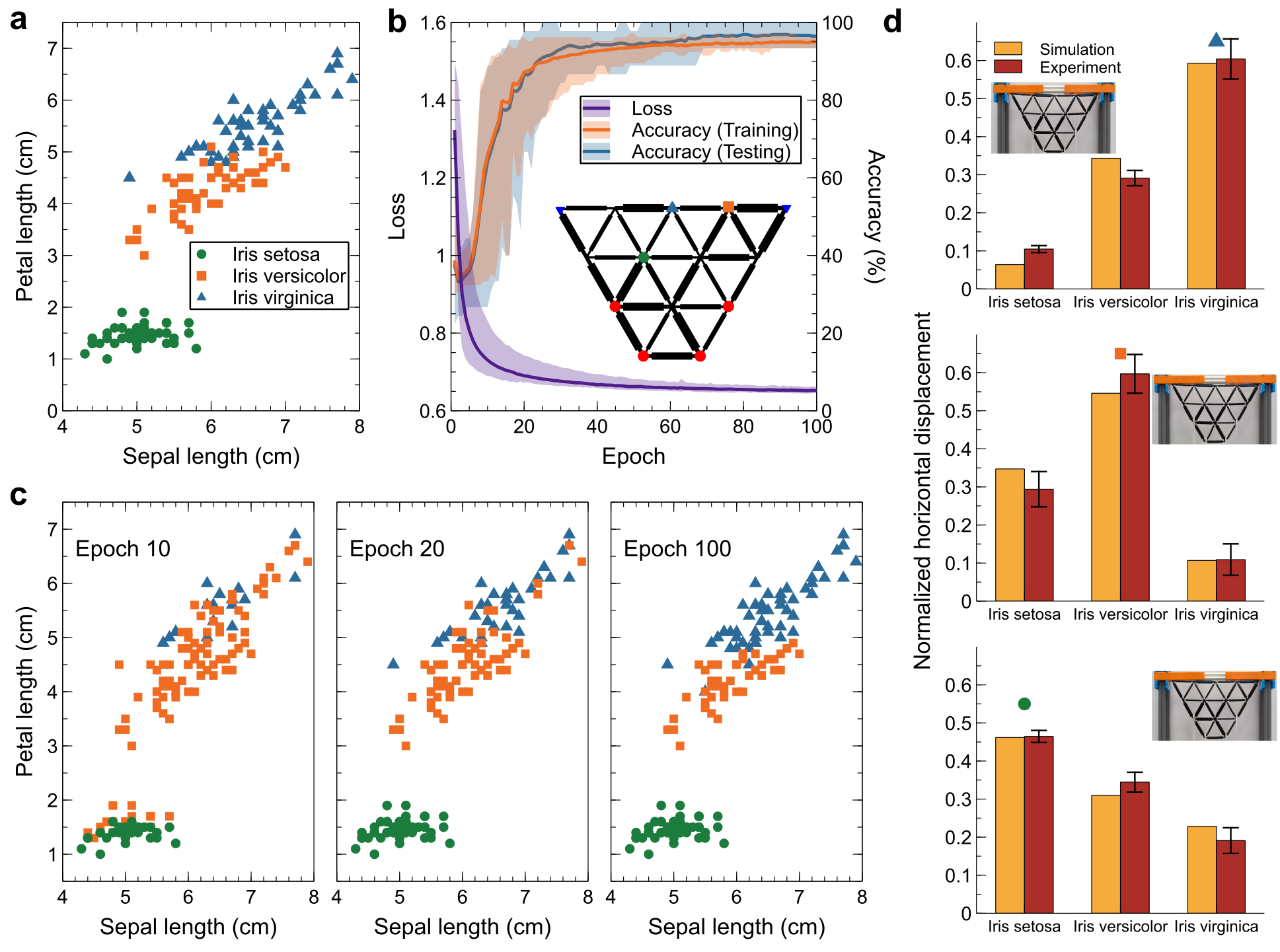}
    \caption{\textbf{Classification using MNNs.}
    \textbf{a} The Iris flower classification dataset. The relation between sepal length and petal length is visualized.
    \textbf{b} The loss~(purple) and classification accuracy~(orange for training set and blue for testing set) as a function of epoch in training process. The inset shows the trained configuration of MNNs. The blue triangles and red dots represent the fixed boundary and the input nodes, respectively. The symbols used in $\textbf{a}$ are shown in the inset of $\textbf{b}$ to represent the output nodes for corresponding type of Iris flowers.
    \textbf{c} The classification results when the epoch is $10$, $20$ and $100$ are shown from left to right, respectively.
    \textbf{d} The comparison of classification results between simulation and experiment when conducted in the MNN at epoch $100$. The insets display the experimental setups. The error bars are calculated based on standard deviation of three independent experiments.
    }
    \label{fig4}
\end{figure}

\begin{figure}[h]
    \centering
    \includegraphics[width=0.9\textwidth]{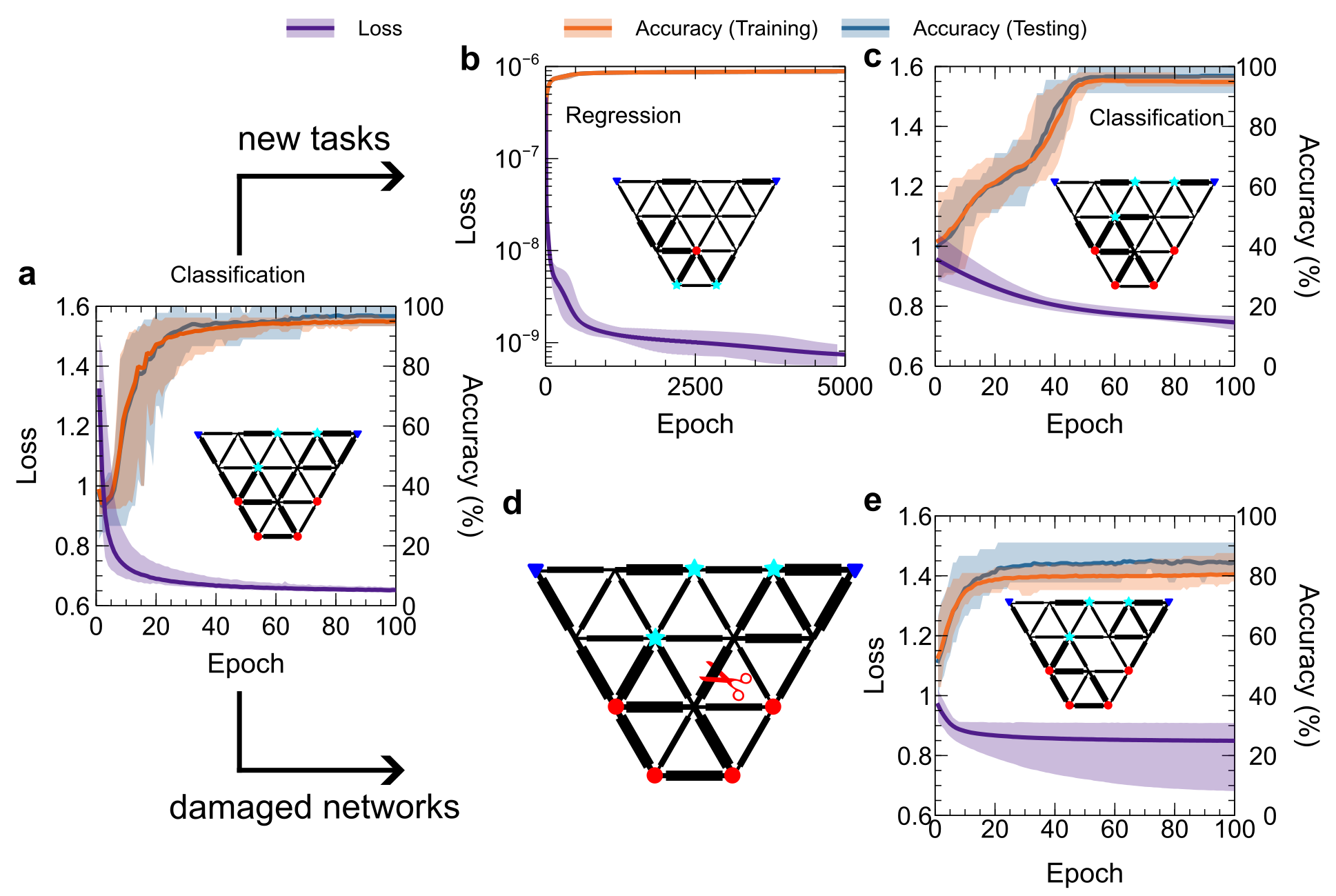}
    \caption{\textbf{Retrainable mechanical networks.}
    \textbf{a} The loss~(purple) and classification accuracy~(orange for training set and blue for testing set) as a function of epoch in the training process of Iris flower classification task. The inset shows the trained MNN.  This MNN is subsequently taken as the initial system for new task training~(top) and retraining after damage~(bottom), respectively.  
    \textbf{b} The loss~(purple) and regression accuracy~(orange) as a function of epoch in the training process when using noise-free dataset and trained MNN of the classification task as an initial MNN. The inset shows the trained MNN.
    \textbf{c} The loss~(purple) and classification accuracy~(orange and blue) as a function of epoch in training process when using trained MNN of the regression task as initial MNN. The inset shows the trained MNN.
    \textbf{d} The schematic shows that a bond of the MNN for classification tasks is pruned.
    \textbf{e} The loss~(purple) and classification accuracy~(orange for training set and blue for testing set) as a function of epoch in training process when using pruned MNN as an initial MNN. The inset shows the trained MNN.
    The blue triangles, red dots and cyan stars in MNNs represent the fixed nodes, the input nodes and the output nodes, respectively.
    }
    \label{fig5}
\end{figure}

It is important to note that the experimental feasibility of in situ backpropagation has been demonstrated as shown in Fig.~\ref{fig1}, where the error signal is backpropagated to each bond to obtain the gradient. Although the update of spring constants is conducted numerically in the current model (Eq.~\eqref{equ7}), the  capabilities of the trained MNNs are validated through experiments. There exist numerous experimental avenues for implementing the update of the spring constant based on the in situ backpropagation we demonstrated here, so that the entire learning process can be implemented experimentally. For example, platforms such as tunable bars~\cite{lee2022mechanical}, and principles such as magnetoactivity~\cite{zhang2023magnetoactive}, phase changing~\cite{poon2019phase}, and phototunability~\cite{stowers2015dynamic}, where material properties can adjust themselves in situ by external fields, hold promise for facilitating further experimentation with in situ backpropagation. Besides, our current in situ backpropagation method is constrained within the linear regime of MNNs, leading to applications primarily focused on linear regression and linear classifier. Therefore, exploring the nonlinear regime of MNNs by using nonlinear materials and geometric nonlinearity presents an opportunity to tackle nonlinear datasets and tasks, thus expanding the capabilities and potential applications of MNNs in the future.

So far, backpropagation has been the most efficient and widely-used neural network training algorithm for machine learning across digital and optical processors~\cite{poggio2020theoretical,lillicrap2020backpropagation,lecun2015deep,wright2022deep,hughes2018training}. Our demonstration of this ubiquitous technique in mechanical systems as a physical implementation unveils the promising capabilities of MNNs to reduce the cost of machine learning. The successful implementation of various tasks using MNNs has wide-ranging implications, bridging the realms of mechanics and machine learning, and paving the way for designing autonomous robots and smart materials with self-learning capabilities, which can not only respond to external stimuli but also possess the ability to learn and adapt to environments.

\section*{Methods}\label{sec5}
\subsection*{Sample fabrication} 
The trained MNN is fabricated using a Polyjet 3D printer~(J850\textsuperscript{TM} Digital Anatomy\textsuperscript{TM}) with Agilus30~(black and flexible). To align with the bar model and prevent the bond-bending force, the joints of bonds in mechanical networks are manufactured thinner~(a half of the bond width), equivalent to half the width of the bonds. This design ensures that the bonds prefer to deform near the nodes in the loading condition, mitigating the risk of buckling in their middle sections.

\subsection*{Experimental setup and measurements}
We use an assembled structure to suspend the MNN by gluing the two nodes~(marked in blue triangles in figures) onto the truss. Thin strings are delicately wound around the joints of the input nodes, serving as hooks to hang weights. Following a careful and slow application of weights onto the strings, experimental procedures involve taking images using a DSLR camera. The parameters for the camera are: F5.0 and ISO800. Positioned on a tripod, the camera is remotely controlled to minimize interference, ensuring precise measurements. The lens is aligned perpendicular to and at the same height as the sample to maintain accuracy. Besides, camera calibration is performed using a standard checkerboard pattern, and images are corrected using the camera matrix. To obtain the elongation of the bonds in mechanical networks, we employ correlation-based algorithms to track the centers of the joints~\cite{hedrick2008software,li2022spatiotemporal,li2023data}. Then, bond elongation is calculated by determining the difference between the original length and the length under applied force. More details can be seen in Supplementary Information.

\subsection*{Numerical simulations}
The simulations of the response of MNNs to applied forces are conducted using bar elements. To verify the feasibility of these simulations in experiments, the finite element analysis of the actual 3D model is also performed on Comsol Multiphysics~(see Supplementary Information). Training of the MNNs is conducted by the in situ backpropagation derived from the adjoint method, as detailed in the main text. The gradient information obtained from this process is utilized to update the spring constant, represented by the width of the bonds, using the Adam optimization algorithm. This process iterates until convergence, achieving a final structure. The initial configurations for the behaviors learning, regression and classification entail MNNs with each bond being $2~\mathrm{mm}$. For experimental purposes, the width of each bond is restricted to range from $1.5~\mathrm{mm}$ to $2.5~\mathrm{mm}$ in the simulation. The learning rate $\alpha$ for the behaviors learning, regression and classification demonstrated in the main text are $0.005$, $0.1$ and $0.006$, respectively. The decay rate for momentum $\beta_{1}$ and the decay rate for squared gradients $\beta_{2}$ are kept to be $0.9$ and $0.999$, respectively. More details can be seen in Supplementary Information.

\backmatter

\bmhead{Supplementary information}
See the attached Supplementary Information.

\bmhead{Acknowledgements}
The authors thank the support from the Office of Naval Research~(MURI N00014-20-1-2479) and National Science Foundation Center for Complex Particle Systems~(Award \#2243104). We are grateful for the fruitful discussions with Xiongye Xiao and Prof. Paul Bogdan at University of Southern California. We also would like to thank Andy Poli in the Department of Mechanical Engineering at the University of Michigan for advise in the fabrication of 3D-printed MNNs.

\section*{Declarations}
\begin{itemize}
\item Funding\\
Office of Naval Research~(MURI N00014-20-1-2479) and National Science Foundation Center for Complex Particle Systems~(Award \#2243104)
\item Competing interests\\
The authors declare no competing interests.
\item Data availability\\
The data in this study are available from the corresponding author on request.
\item Code availability\\
The codes in this study are available from the corresponding author on request.
\item Author contribution\\
S.L. and X.M. designed the project. S.L. conducted theoretical derivation, numerical simulations and experiments. S.L. and X.M. wrote and improved the manuscript.
\end{itemize}

\clearpage
\bibliography{sn-bibliography}

\end{document}